\documentclass[a4paper,twoside]{article}

\usepackage{epsfig}
\usepackage{subcaption}
\usepackage{calc}
\usepackage{amssymb}
\usepackage{amstext}
\usepackage{amsmath}
\usepackage{amsthm}
\usepackage{multicol}
\usepackage{pslatex}
\usepackage{apalike}
\usepackage{algorithm2e}
\usepackage[bottom]{footmisc}

\usepackage{multirow}
\usepackage{booktabs}
\usepackage{diagbox}
\usepackage{url}
\usepackage{hyperref}

\usepackage{SCITEPRESS}     

\begin{document}

\title{ZQBA: Zero Query Black-box Adversarial Attack}

\author{\authorname{Joana C. Costa\sup{1}\orcidAuthor{0000-0003-3995-4621}, Tiago Roxo\sup{1}\orcidAuthor{0000-0001-9563-8039}, Hugo Proença\sup{1}\orcidAuthor{0000-0003-2551-8570} and Pedro R. M. Inácio\sup{1}\orcidAuthor{0000-0001-8221-0666}}
\affiliation{\sup{1}sins-lab, Instituto de Telecomunicações, Universidade da Beira Interior, R. Marquês D'Ávila e Bolama, Covilhã, Portugal}
\email{\{joana.cabral.costa, tiago.roxo, hugomcp, prmi\}@ubi.pt}
}


\abstract{Current black-box adversarial attacks either require multiple queries or diffusion models to produce adversarial samples that can impair the target model performance. However, these methods require training a surrogate loss or diffusion models to produce adversarial samples, which limits their applicability in real-world settings. Thus, we propose a Zero Query Black-box Adversarial (ZQBA) attack that exploits the representations of Deep Neural Networks (DNNs) to fool other networks. Instead of requiring thousands of queries to produce deceiving adversarial samples, we use the feature maps obtained from a DNN and add them to clean images to impair the classification of a target model. The results suggest that ZQBA can transfer the adversarial samples to different models and across various datasets, namely CIFAR and Tiny ImageNet. The experiments also show that ZQBA is more effective than state-of-the-art black-box attacks with a single query, while maintaining the imperceptibility of perturbations, evaluated both quantitatively (SSIM) and qualitatively, emphasizing the vulnerabilities of employing DNNs in real-world contexts. All the source code is available at~\url{https://github.com/Joana-Cabral/ZQBA}.}

\onecolumn \maketitle \normalsize \setcounter{footnote}{0} \vfill

\section{\uppercase{Introduction}}

Deep Neural Networks (DNNs) have achieved excellent performance in multiple areas, such as Medical Imaging~\cite{thirunavukarasu2023large,patricio2023coherent}, Natural Language Processing~\cite{touvron2023llama2openfoundation,costa2022predicting}, and Active Speaker Detection~\cite{roxo2024bias,roxo2024asdnb}, which influenced a widespread approval of using Artificial Intelligence in the daily routine. This increases the appeal for attackers to explore vulnerabilities in DNNs, and, as highlighted by Szegedy \textit{et al.}~\cite{szegedy2014intriguing}, these models fail to generalize and are vulnerable to noise imperceptible to the human eye, meaning that they are susceptible to adversarial attacks. In this context, there are two key approaches: \textit{white-box}, which assumes that the attacker can access the gradients of the target model, and \textit{black-box}, which considers that the attacker can not access the gradients, but can query the target model multiple times. Recent works regarding \textit{black-box} scenarios use only input images and predictions to estimate the gradients~\cite{chen2017zoo} of a target model or train generative models~\cite{liu2024difattack,chen2024diffusion} to create strong perturbations. However, these approaches are either dependent on thousands of queries to the target model or training generative models on adversarial samples, which may not be feasible in real-world settings.

This paper proposes Zero Query Black-box Adversarial (ZQBA) attack, which creates adversarial perturbations using feature maps extracted from various DNNs to fool other networks (target models), without requiring queries or training additional generative models. These feature maps are then combined with the clean images to reduce the target model performance, without impairing image quality, which is evaluated using Structure Similarity (SSIM)~\cite{wang2004image} that measures the quality of an image according to the perception of the human eye. The results indicate that the generated adversarial samples have transferability between various models and across different datasets, and is competitive with state-of-the-art black-box attacks using a single query. This paper contributions can thus be summarized as follows:

\begin{itemize}
    \item We propose ZQBA, an adversarial black-box attack that does not require querying the target model thousands of times nor relies on training a generative model to produce adversarial samples;
    
    \item We present an attack strategy that leverages the feature map representations from DNNs, distinctive from the target models, by adding these maps to the clean images to produce adversarial samples without significant overhead;
    
    \item Ablation studies and experimental evaluation demonstrate ZQBA is a competitive approach relative to state-of-the-art black-box using a single query, does not impair the image quality perception, and has high transferability between models and across different datasets.
\end{itemize}

The remaining of the paper is structured as follows: section~\ref{sec:rel_work} discusses the related works; section~\ref{sec:methodology} describes the attack scenario and provides an overview of the ZQBA strategy; section~\ref{sec:experiments} reports the experimental setup, ablation studies, and performance analysis, accompanied by a discussion; finally, Section~\ref{sec:conclusions} concludes the paper.

\section{\uppercase{Related Work}}
\label{sec:rel_work}

\textbf{White-box Attacks} rely on assessing model gradients to create adversarial samples.
Fast Gradient Sign Method (FGSM)~\cite{goodfellow2014explaining} is a one-step method that finds adversarial samples using the model loss function.
Jacobian-based Saliency Maps (JSM)~\cite{papernot2016limitations} directly maps the input to the output and only alters small fractions of the image.
DeepFool~\cite{moosavi2016deepfool} uses iterative linearization to create minimal perturbations that cross the decision boundary.
Projected Gradient Descent (PGD)~\cite{madry2018towards} uses saddle point formulation to find a strong perturbation through multiple iterations.
SmoothFool~\cite{dabouei2020smoothfool}, uses smooth adversarial perturbations to enhance transferability and stealthiness.
Auto-Attack~\cite{croce2020reliable} combines four attacks (the majority are white-box) to evaluate the robustness of a defense. The need of white-box attacks to access the model gradients to create adversarial samples limits its applicability in real-world scenarios, diverging from ZQBA scope.

\textbf{Multiple Query Black-box Attacks} require thousands of queries to the target models to create a suitable adversarial sample.
Zeroth Order Optimization (ZOO)~\cite{chen2017zoo} directly estimates the gradients of the model without training an additional one, achieving better transferability.
Square Attack~\cite{andriushchenko2020square} uses a randomized search scheme that perturbs the images in localized square-shaped regions in random positions.
Sparse Adversarial and Imperceptible Attack~\cite{imtiaz2022saif} uses conditional gradient to optimize attack perturbations, maintaining low magnitude and sparsity.
Park \textit{et al.}~\cite{park2024hard} propose a practical way of using hard-label-based attacks, using a surrogate model, to achieve higher query efficiency. Contrary to these approaches, we generate attacks that do not require interaction with the target model and generalize to other models.

\textbf{Diffusion Black-box Attacks} train diffusion models to produce adversarial samples.
Chen \textit{et al.}~\cite{chen2023content} propose an unrestricted attack framework by mapping adversarial samples to a low-dimensional manifold.
AdvDiffuser~\cite{chen2023advdiffuser} manipulates the latent code by using class activation maps to perturb less important areas.
DiffAttack~\cite{chen2024diffusion} influences the latent space to generate human-insensitive perturbations with semantic indications.
TAIGen~\cite{roy2025taigen} uses unconditional diffusion models and employs a selective RGB channel strategy.
ScoreAdv~\cite{huang2025scoreadv} incorporates an interpretable saliency map in the diffusion model to produce realistic adversarial images. Contrary to these approaches, ZQBA uses models trained for classification and does not rely on diffusion models to create adversarial samples.

\begin{figure*}[!tb]
    \centering
    \includegraphics[width=0.8\textwidth]{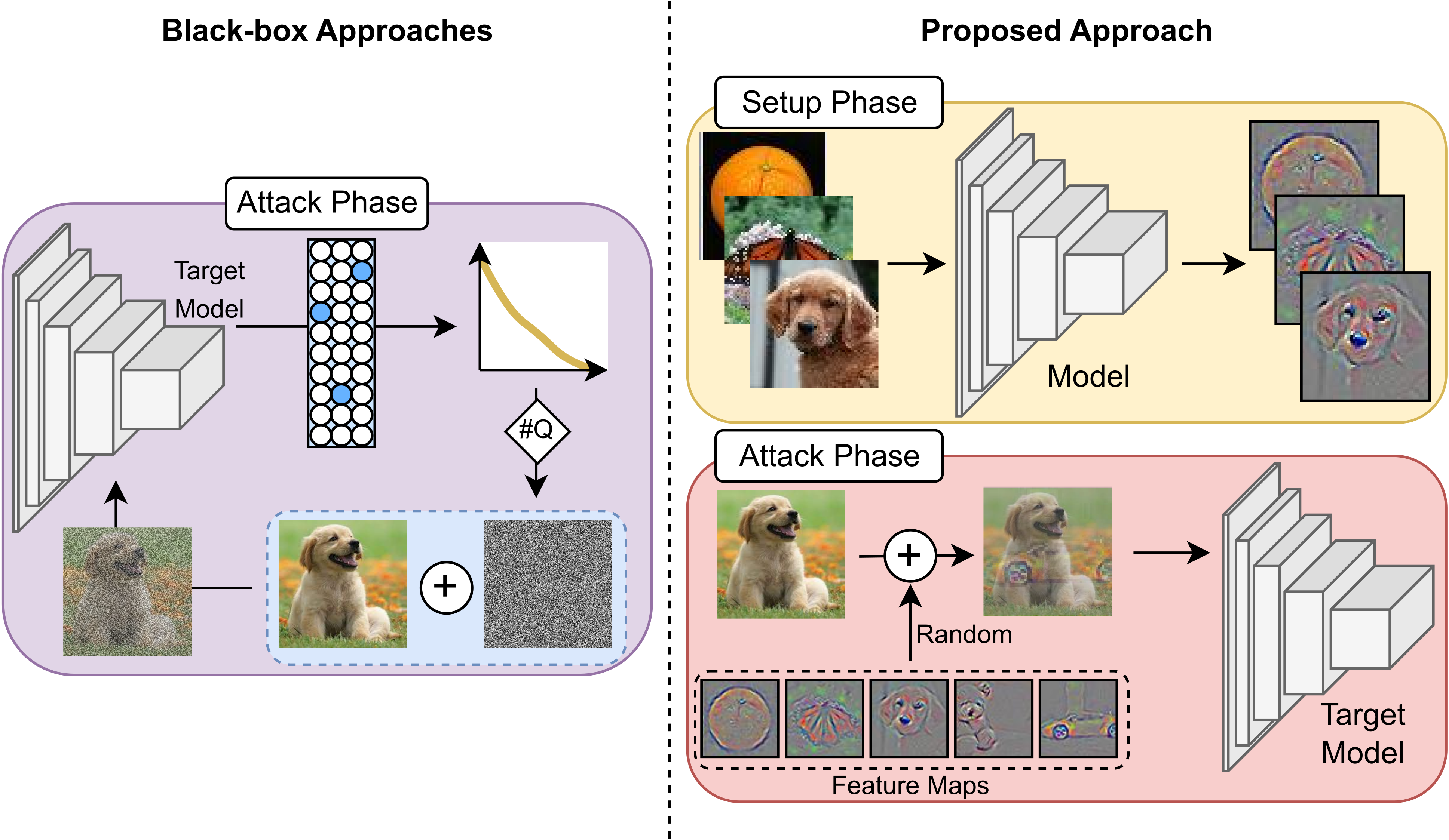}
    \caption{Comparison between multiple queries black-box approaches and the proposed attack: Zero-Query Black-box Adversarial (ZQBA) attack. During the attack phase, black-box attacks query the target model thousands of times to obtain the logits related to the provided images, and adapt a loss function based on the model responses to generate better perturbations. On the other hand, ZQBA has two phases: 1) Setup, where all the feature maps (perturbations) are obtained; and 2) Attack, where previously obtained perturbations are added to the image to be attacked, without querying the target model.}
    \label{fig:black_box_vs_proposed}
\end{figure*}

\section{\uppercase{Preliminaries and Methodology}}
\label{sec:methodology}

\subsection{Attack Scenario}

This paper assumes that an attacker is trying to impair the accuracy of a Deep Learning model through adversarial attacks. Depending on the amount of information, there are two types of attacks: 1) \textit{white-box}, which require access to the gradients of the target model; and 2) \textit{black-box}, which demand multiple queries to the target model to simulate its behavior. However, black-box can be expensive or non-feasible in short periods of time due to thousands of queries, suggesting more realistic attack scenarios~\cite{costa2025lisard}. In this case, \textbf{we propose creating adversarial attacks without querying the target model} by adding the representation of a model different from the target one, while balancing the imperceptibility and effectiveness of the perturbation. Figure~\ref{fig:black_box_vs_proposed} illustrates the difference between the proposed and current black-box scenarios, demonstrating that ZQBA does not require multiple iterations to create the perturbation, highlighting that it can impair the accuracy of the target model without needing to know the model behavior.

\subsection{Image Perturbation for Attacks} 

\begin{table}[!tb]
    \tiny
    \centering
    \caption{Performance of multiple architectures solely on clean examples on different datasets. Par (M) refers to the number of parameters in millions.}
    \begin{tabular}{c|cccc}
        \toprule
        \multirow{2}{*}{\textbf{Model}} & \multirow{2}{*}{\textbf{Par (M)}} & \multicolumn{3}{c}{\textbf{Clean Accuracy}} \\
        && CIFAR-10 & CIFAR-100 & TinyImageNet \\
        \toprule
        MobileNetv2 & 3.3 & 85.08 & 57.27 & 45.79 \\
        EfficientNetB2 & 8.7 & 84.99 & 58.70 & 58.90 \\
        ResNet18 & 11.1 & 94.43 & 69.64 & 55.85 \\
        ResNet50 & 24.4 & \textbf{96.65} & 78.33 & 63.83 \\
        WideRN28-10 & 36.4 & 89.52 & 56.43 & 48.11 \\
        ResNet101 & 42.5 & 96.25 & \textbf{78.64} & \textbf{63.99} \\
        VGG19 & 137.0 & 91.63 & 63.17 & 55.38 \\
        \bottomrule
    \end{tabular}
    \label{tab:parameters_clean_accuracy}
\end{table}

\begin{figure}
    \centering
    \includegraphics[width=0.4\textwidth]{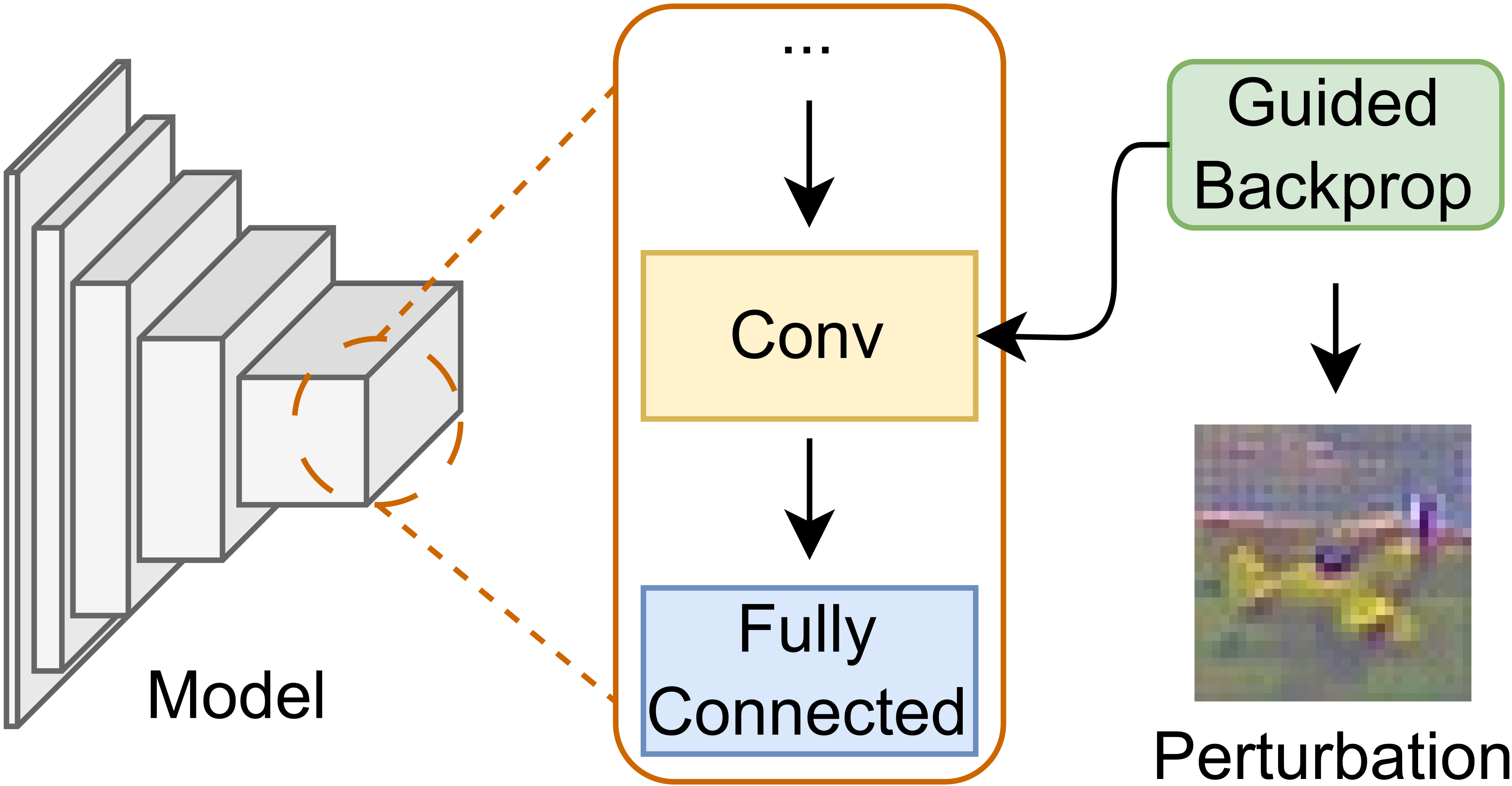}
    \caption{Overview of the methodology used to obtain the perturbations for the ZQBA attack, through the extraction of feature maps, using Guided Backpropagation, from the layer prior to the classification layers.}
    \label{fig:guided_backprop}
\end{figure}

\textbf{Obtaining Feature Maps.} 
Our approach involves training a model and obtaining its representations for different classes by extracting feature maps from layers prior to the classification ones (\textit{i.e.}, fully connected layers) using backpropagation, as shown in Figure~\ref{fig:guided_backprop}. To obtain these feature maps, we use Guided Backpropagation~\cite{mostafa2022leveraging}, which is a gradient-based approach that retrieves the gradient of images when backpropagating through the Rectified Linear Unit (ReLU) activation function, where only the flow of positive gradients is allowed by changing the negative gradient values to zero. This approach is typically used to display the features of the input image that maximize the activation of the feature maps, resembling the features that more closely influence the model prediction. Other possible approaches to obtain the visual representation of the models could have GradCAM and its variants. However, Guided Backpropagation demonstrated greater potential for applicability since it clearly represented the object, rather than attention areas of the model.

\textbf{Creating Adversarial Images.}
The idea behind ZQBA is to maintain the original image representation while adding perturbations that do not significantly alter this image, while impairing the classification of a model. Therefore, after obtaining the feature maps for every image in a dataset, ZQBA conjugates a randomly selected feature map with a clean image, according to the following equation:
\begin{equation}
\textbf{X}_{adv} = \alpha \cdot (\nabla_\textbf{X} \cdot f(\textbf{X})) + \textbf{X},
\end{equation}
where $\textbf{X}_{adv}$ is the perturbed image, $\alpha$ is the perturbation weight, $\nabla_\textbf{X}$ is the gradient with respect to the input image pixels, $f(\textbf{X})$ is the logits corresponding to the model predicted class, and $\textbf{X}$ is the original image. ZQBA considers $\alpha = 0.4$ since it achieved the best results in performance and image quality, as shown in the Ablation Studies (section~\ref{sec:abl_studies}). 

\textbf{Feature Map Selection.} The selection of the feature maps could either be based on the similarity between the clean image and the image used to obtain the feature map, or by choosing a random feature map, which ultimately increases the viability of the attack and its application range, as shown in the Ablation Studies (section~\ref{sec:abl_studies}). Usually, when perturbations are applied to an image, there is a maximum threshold to limit the amount of manipulation in the clean image ($\epsilon=8/255$ for CIFAR-10 and CIFAR-100 and $\epsilon=4/255$ for Tiny ImageNet). In the ZQBA context, this threshold does not directly apply since it is used to limit the amount of noise generated based on the gradients of the model for \textit{white-box} or based on the number of queries for \textit{black-box}. To ensure that the attacked images created by ZQBA remain imperceptible, we assess the quality of the generated images using Structural Similarity (SSIM)~\cite{wang2004image}, which evaluates the images according to the perception of the human eye, and ensure that SSIM remains above or equal to 0.95 (SSIM $>=0.95$), at which the human observer can not detect manipulations to the image~\cite{flynn2013image}.

\begin{table*}[!tb]
    \small
    \centering
    \caption{Performance of ZQBA in cross-architecture settings in different datasets. MNv2, E2, RN18, RN50, RN101, and WRN refer to MobileNetv2, EfficientNetB2, ResNet18, ResNet50, ResNet101, and WideResNet28-10, respectively. F and T refer to the feature maps and the target model, respectively. $\Delta$ displays the difference between clean and strongest attack (bold) accuracies.}
    \begin{tabular}{c|c|c|c|c|c|c|c|c|c}
        \toprule
        \textbf{Dataset} & \diagbox{T}{F} & MNv2 & E2 & RN18 & RN50 & WRN & RN101 & VGG19 & $\Delta$ \\
        \toprule
        \multirow{7}{*}{CIFAR-10} &
        MNv2 & \textbf{50.39} & \underline{54.22} & 67.27 & 69.18 & 65.35 & 69.91 & 62.43 & -34.69 \\
        &
        E2 & \textbf{53.69} & \underline{54.39} & 67.28 & 68.98 & 65.50 & 69.41 & 62.88 & -31.30 \\
        &
        RN18 & \textbf{73.07} & 74.73 & 78.63 & 81.17 & \underline{74.18} & 81.22 & 75.24 & -21.36 \\
        &
        RN50 & \textbf{76.22} & \underline{77.19} & 81.33 & 82.00 & 80.26 & 83.50 & 78.61 & -20.43 \\
        &
        WRN & \textbf{65.57} & \underline{67.05} & 71.83 & 74.98 & 68.71 & 74.79 & 69.52 & -23.95 \\
        &
        RN101 & \textbf{75.33} & 77.28 & 79.05 & 79.73 & 78.66 & 82.53 & \underline{76.04} & -20.92 \\
        &
        VGG19 & \textbf{66.51} & \underline{67.42} & 74.00 & 75.55 & 71.71 & 76.43 & 67.50 & -25.12 \\ 
        \midrule
        
        \multirow{7}{*}{CIFAR-100} &
        MNv2 & \textbf{16.36} & \underline{22.02} & 30.07 & 34.72 & 33.41 & 33.45 & 23.17 & -40.91 \\
        &
        E2 & \textbf{25.82} & \underline{25.87} & 32.98 & 35.34 & 35.43 & 35.07 & 28.53 & -32.88 \\
        &
        RN18 & \textbf{30.18} & 34.20 & 42.16 & 45.29 & 46.86 & 42.31 & \underline{32.16} & -39.46 \\
        &
        RN50 & \textbf{38.98} & 41.27 & 46.84 & 51.20 & 54.15 & 53.02 & \underline{40.91} & -39.35 \\
        &
        WRN & 36.31 & 36.24 & 34.88 & 36.47 & \textbf{29.80} & 37.38 & \underline{33.98} & -26.63 \\
        &
        RN101 & \underline{39.27} & 42.42 & 47.13 & 53.19 & 54.95 & 51.51 & \textbf{38.23} & -40.41 \\
        &
        VGG19 & \underline{30.88} & 31.08 & 37.44 & 41.13 & 38.46 & 39.87 & \textbf{27.19} & -35.98 \\ 
        \midrule
        
        \multirow{7}{*}{Tiny ImageNet} &
        MNv2 & \textbf{16.23} & \underline{22.46} & 25.39 & 26.56 & 26.21 & 25.09 & 25.51 & -29.56 \\
        &
        E2 & 37.92 & \textbf{35.09} & 37.13 & 38.69 & \underline{37.12} & 37.36 & 39.57 & -23.81 \\
        &
        RN18 & 36.38 & 36.03 & \textbf{29.43} & 34.35 & 34.14 & \underline{32.63} & 35.90 & -26.42 \\ 
        &
        RN50 & 44.49 & 43.31 & 38.59 & \underline{37.72} & 41.60 & \textbf{37.09} & 43.53 & -26.74 \\ 
        &
        WRN & 30.58 & 30.87 & \underline{29.52} & 31.92 & \textbf{27.86} & 32.32 & 31.90 & -20.25 \\
        &
        RN101 & 42.73 & 41.29 & \underline{38.06} & 39.46 & 41.62 & \textbf{33.06} & 42.41 & -30.93 \\ 
        &
        VGG19 & 34.09 & 34.83 & \textbf{33.52} & 36.01 & 35.41 & 35.54 & \underline{33.54} & -21.86 \\ 
        \bottomrule
    \end{tabular}
    \label{tab:cross_architecture}
\end{table*}

\section{\uppercase{Experiments}}
\label{sec:experiments}

\subsection{Experimental Setup}

\textbf{Datasets.} The models are evaluated on CIFAR10~\cite{krizhevsky2009learning} and CIFAR100~\cite{krizhevsky2009learning}, consisting of 50,000 training and 10,000 testing images, and Tiny ImageNet~\cite{le2015tiny}, which has 100,000 training and 10,000 validation images, and is a subset of ImageNet comprising only 200 classes.

\textbf{Models.} Each selected architecture was trained in the CIFAR10, CIFAR100, and Tiny ImageNet datasets, and models with the best \textit{clean accuracy} were selected. These models are used to obtain the feature maps for each dataset, and the models susceptible to perturbations (target models) differ from those used to obtain these feature maps. Similar to the literature, the first Convolutional layer of ResNet18, 50, and 101 was modified to be more suitable for the image size of CIFAR-10 and CIFAR-100. The remaining networks follow the implementation provided by PyTorch~\cite{paszke2019pytorch}.

\textbf{Evaluation}. We use accuracy on both natural test samples, denoted as \textit{Clean Accuracy}, and to assess the performance o f the models to ZQBA adversarial samples. We evaluate the quality of the attacked images using SSIM~\cite{wang2004image} metric, relating to human eye perception.

\subsection{ZQBA Performance Evaluation}

The achieved clean accuracy for the different datasets and architectures is displayed in Table~\ref{tab:parameters_clean_accuracy}, with the respective number of parameters. Although it was expected that bigger networks would achieve greater clean accuracy, the ResNet family ended up in the top 3 results for CIFAR-10 and CIFAR-100 due to the change in the first Convolutional layer to be adapted to a smaller image size. These networks will be utilized in the remaining experiments.

\begin{table}[!tb]
    \tiny
    \centering
    \caption{Transferability of ZQBA in multiple datasets. The top row in the header refers to the dataset used to obtain the feature maps, and the last row in the header refers to the evaluated dataset. The value between parentheses refers to the decrease relative to clean accuracy.}
    \begin{tabular}{c|cccc}
        \toprule
        \multirow{3}{*}{\textbf{Model}} 
        & CIFAR-10 & CIFAR-100 & Tiny & Tiny \\
        & $\downarrow$ & $\downarrow$ & $\downarrow$ & $\downarrow$ \\
        & Tiny & Tiny & CIFAR-10 & CIFAR-100 \\
        \toprule
        MNv2 & 27.60 (40\%) & 27.91 (39\%) & 78.00 (8\%) & 42.54 (26\%) \\
        E2 & 40.34 (32\%) & 40.55 (31\%) & 77.12 (9\%) & 44.90 (24\%) \\
        RN18 & 38.78 (43\%) & 38.99 (43\%) & 89.39 (5\%) & 55.10 (21\%) \\
        RN50 & 47.74 (39\%) & 49.21 (37\%) & 93.04 (4\%) & 66.79 (15\%) \\
        WRN & 31.61 (34\%) & 29.30 (39\%) & 81.80 (9\%) & 36.96 (35\%) \\
        RN101 & 49.51 (35\%) & 47.41 (38\%) & 92.43 (4\%) & 66.83 (15\%) \\
        VGG19 & 38.37 (31\%) & 37.46 (32\%) & 86.74 (5\%) & 54.18 (14\%) \\
        \bottomrule
    \end{tabular}
    \label{tab:cross_domain}
\end{table}

\textbf{Architecture Influence on Attack Performance.}
The first experiment focuses on evaluating the performance of ZQBA in different architectures and datasets to determine whether the feature maps generalize across various architectures or are dependent on prior knowledge. Table~\ref{tab:cross_architecture} displays the accuracy under ZQBA attack for cross-architecture in CIFAR-10, CIFAR-100, and Tiny ImageNet. The obtained results show that: 1) for CIFAR-10, the smallest networks generate the best perturbations for all architectures; 2) for CIFAR-100, the smallest network achieves the best overall results, except for the larger networks, where the same architecture generates better perturbations; and 3) for Tiny ImageNet, overall the same network type is the best at generating the perturbations. In sum, the feature maps used to generate adversarial images are not dependent on the architecture of the target model (\textit{i.e.}, the performance of the models is always below the clean accuracy presented in Table~\ref{tab:parameters_clean_accuracy}, as highlighted by the $\Delta$), meaning that the attacker does not need previous details of the target model architecture, although this knowledge may increase the strength of the attack.

\textbf{Dataset Transferability.}
We evaluate the transferability of ZQBA to datasets distinct from the one used to generate the feature maps, in Table~\ref{tab:cross_domain}, highlighting that models used to obtain these maps do not need to be trained on the same data as the target models. Since CIFAR-10 and CIFAR-100 share the same data, we do not consider this combination in this experiment. The transferability strength is evaluated using the model accuracy and its relative decrease to the clean accuracy, represented by the values between parentheses, as a percentage. The results show that: 1) the transferability from CIFAR-10/100 to Tiny ImageNet has a significant impact on the model accuracy, justified by the perturbations having a smaller size with a more scattered representation of the objects in the image; and 2) the perturbations obtained from Tiny ImageNet have less impact due to its size being adjusted to match the image size of the target dataset, resulting in a loss of detail in the used perturbation; and 3) the difference between Tiny$\rightarrow$CIFAR10 and Tiny$\rightarrow$CIFAR100, which relates to the inherent complexity of manipulating the classifications for a dataset with fewer classes (10 \textit{vs.} 100), \textit{i.e.}, perturbations are more likely to impair the classification of a model with 100 classes than one with 10 classes, given that the latter has more precise decision boundaries for classification. These results indicate that transferability of ZQBA is improved when perturbations are obtained from smaller images and applied to larger images, and that datasets with more classes are more susceptible to changes in final classification.

\begin{table}[!tb]
    \tiny
    \centering
    \caption{Performance of ZQBA and state-of-the-art black-box attacks, using ResNet model, in different datasets.}
    \begin{tabular}{c|ccccc}
        \toprule
        \textbf{Attack} & \textbf{\#Queries} & \textbf{CIFAR-10} & \textbf{CIFAR-100} & \textbf{Tiny ImageNet}\\
        \toprule
        Square & 1 & 88.47 & 51.55 & 44.03 \\
        ZOO & 1 & 94.43 & 69.63 & 67.78 \\ 
        \midrule
        \textbf{ZQBA} & 0 & \textbf{78.63} & \textbf{42.16} & \textbf{29.43} \\
        \bottomrule
    \end{tabular}
    \label{tab:SOTA_comparison}
\end{table}

\textbf{State-of-the-art Comparison.} Current black-box attacks use multiple queries to obtain perturbations for the target models, while our approach does not require any previous interaction with these models. For a fair comparison with ZQBA, we consider state-of-the-art black-box adversarial attacks with a single query. Table~\ref{tab:SOTA_comparison} presents the results for a ResNet-18, as the reference model, on various datasets. Regarding the ZOO attack, the accuracy remains the same as the clean accuracy since the attack can not, in one query, reliably find a perturbation. Furthermore, Square attack shows a slight decrease in model performance, but its efficiency relies on a greater number of queries due to its approach of refining the noise added to the image. ZQBA has a greater performance decrease relative to the clean accuracy, showing the impact of this attack without queries for all the considered datasets.

\begin{table}[!tb]
    \tiny
    \centering
    \caption{Performance ZQBA using a ResNet model when susceptible to multiple variations of noise, in different datasets. SSIM is used to assess image quality.}
    \begin{tabular}{@{}c|c@{}cc@{}cc@{}c}
        \toprule
        \multirow{2}{*}{\textbf{Variations}} & \multicolumn{2}{c}{\textbf{CIFAR-10}} & \multicolumn{2}{c}{\textbf{CIFAR-100}} & \multicolumn{2}{c}{\textbf{Tiny ImageNet}} \\
         & Acc. & SSIM ($\uparrow$) & Acc. & SSIM ($\uparrow$) & Acc. & SSIM ($\uparrow$) \\
        \toprule
        Feature Map & \textbf{78.63} & \textbf{0.95} & \textbf{42.16} & \textbf{0.95} & \textbf{29.43} & \textbf{0.95} \\
        EigenGradCAM & 90.40 & 0.83 & 58.50 & 0.88 & 43.38 & 0.89 \\
        GradCAM & 90.58 & 0.83 & 58.61 & 0.88 & 43.39 & 0.89 \\
        FullGrad & 91.77 & 0.86 & 60.89 & 0.89 & 46.43 & 0.94 \\
        \bottomrule
    \end{tabular}
    \label{tab:extraction_methods_variations}
\end{table}

\begin{table}[!tb]
    \tiny
    \centering
    \caption{Performance of ZQBA using a ResNet model when selecting the feature maps based on different image similarity metrics, in different datasets. SSIM is used to assess image quality.}
    \begin{tabular}{@{}c|cl@{}cl@{}cl@{}c}
        \toprule
        \multirow{2}{*}{\textbf{Approach}} & \multirow{2}{*}{\textbf{Metric}} & \multicolumn{2}{c}{\textbf{CIFAR-10}} & \multicolumn{2}{c}{\textbf{CIFAR-100}} & \multicolumn{2}{c}{\textbf{Tiny ImageNet}} \\
         & & Acc. & SSIM($\uparrow$) & Acc. & SSIM($\uparrow$) & Acc. & SSIM($\uparrow$) \\
        \toprule
        \multirow{3}{*}{Least} 
        & RMSE & 78.46 & 0.95 & 41.97 & 0.95 & 31.28 & 0.95 \\
        & SSIM & 83.80 & 0.95 & 48.32 & 0.95 & 30.33 & 0.95 \\
        & Pearson & 78.68 & 0.95 & 42.30 & 0.95 & 31.24 & 0.95 \\
        \midrule
        \multirow{3}{*}{Most} 
        & RMSE & 83.09 & 0.95 & 46.93 & 0.95 & 33.17 & 0.95 \\
        & SSIM & 78.66 & 0.95 & 41.80 & 0.95 & 30.49 & 0.95 \\
        & Pearson & 79.19 & 0.95 & 42.65 & 0.95 & 30.69 & 0.95 \\
        \midrule
        Random & - & 78.63 & 0.95 & 42.16 & 0.95 & 29.43 & 0.95 \\
        \bottomrule
    \end{tabular}
    \label{tab:approach_similarity_variants}
\end{table}

\subsection{Ablation Studies}
\label{sec:abl_studies}

\textbf{Selecting Noise Perturbation.}
The motivation behind ZQBA is to use information obtained from DNNs to create effective perturbations that deceive model classification. Thus, we explore various methods for extracting the representative information of the original images, using Feature Maps, EigenGradCAM~\cite{muhammad2020eigen}, GradCAM~\cite{selvaraju2017grad}, and FullGrad~\cite{srinivas2019full} in Table~\ref{tab:extraction_methods_variations}. The results show that extracting the Feature Maps, using Guided Backpropagation, conducts to greater performance decrease and achieves better image quality, as depicted by the SSIM values, throughout the datasets. Furthermore, the FullGrad approach has less effective results (greater accuracy), despite achieving better image quality than other GradCAM approaches due to more concise attention maps. In sum, the Feature Maps provide the best relation between image quality and performance decrease given its representation of the entire object in the image, as opposed to the GradCAM approaches, which emphasize specific parts of the image (\textit{i.e.}, model attention).

\textbf{Image Similarity Effect.}
We further assess if the attack, using feature maps, would have greater impact based on the similarity between the images used to obtain the feature maps and the target image, in Table~\ref{tab:approach_similarity_variants}. We do not consider feature maps from the same class as the target image, since it would lead the model to correctly classify the attacked image. The results show that selecting the feature map based on similarity does not affect the quality of the attacked image, since all approaches achieved adequate image quality ($>= 0.95$). Regarding the accuracy, randomly selecting feature maps is a reliable strategy for all the datasets and has the inherent advantage of not requiring a preliminary evaluation of the similarity between images, with increasing overhead with dataset size. Therefore, the considered strategy was the Random selection of feature maps to perturb the image, which further diminishes the knowledge required by the attacker, widening the applicability of ZQBA.

\begin{figure}
\centering
    \begin{subfigure}{.4\textwidth}
      \centering
      \includegraphics[width=\textwidth]{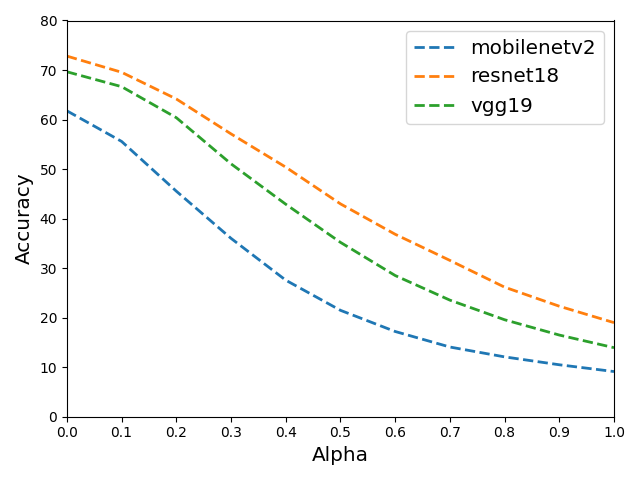}
      \caption{}
      \label{fig:alpha_accuracy}
    \end{subfigure}
    \begin{subfigure}{.4\textwidth}
      \centering
      \includegraphics[width=\textwidth]{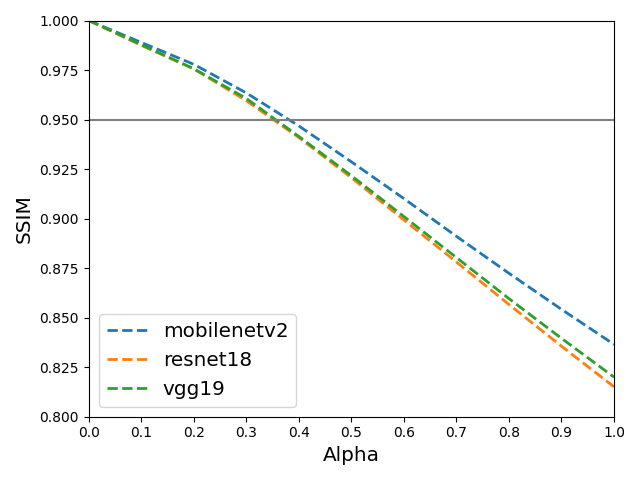}
      \caption{}
      \label{fig:alpha_ssim}
    \end{subfigure}
\caption{Accuracy (\%) and SSIM (\%) of ZQBA with different weights for the feature maps in multiple architectures. The lines refer to an average of each model performance across the three considered datasets.}
\label{fig:alpha_variations}
\end{figure}

\textbf{Effect of Feature Map Weight.}
When adding the perturbations to the target images, ZQBA considers an $\alpha$ value to weight the impact of the perturbation relative to the original image that decreases models performance, without compromising image quality, \textit{i.e.} SSIM $>= 0.95$~\cite{flynn2013image}. We evaluate different $\alpha$ values, in Figure~\ref{fig:alpha_variations}, relative to accuracy and SSIM, considering different architectures, with the values averaged through the three datasets. We choose MobileNetv2, ResNet18, and VGG19 to depict diverse model families and sizes. The results show that increasing the value of $\alpha$ reduces model accuracy, regardless of the architecture, as shown in Figure~\ref{fig:alpha_accuracy}. However, solely considering accuracy to assess the best value for $\alpha$ is not sufficient since ZQBA needs to produce perturbations imperceptible to the human eye. As such, Figure~\ref{fig:alpha_ssim} displays the relation between $\alpha$ and SSIM, with the threshold for image quality ($0.95$), and the results show a decrease in image quality with the increase of $\alpha$, emphasizing that the best value above the threshold is $0.4$, throughout the architectures and datasets.

\begin{figure}[!tb]
    \centering
    \includegraphics[width=0.8\linewidth]{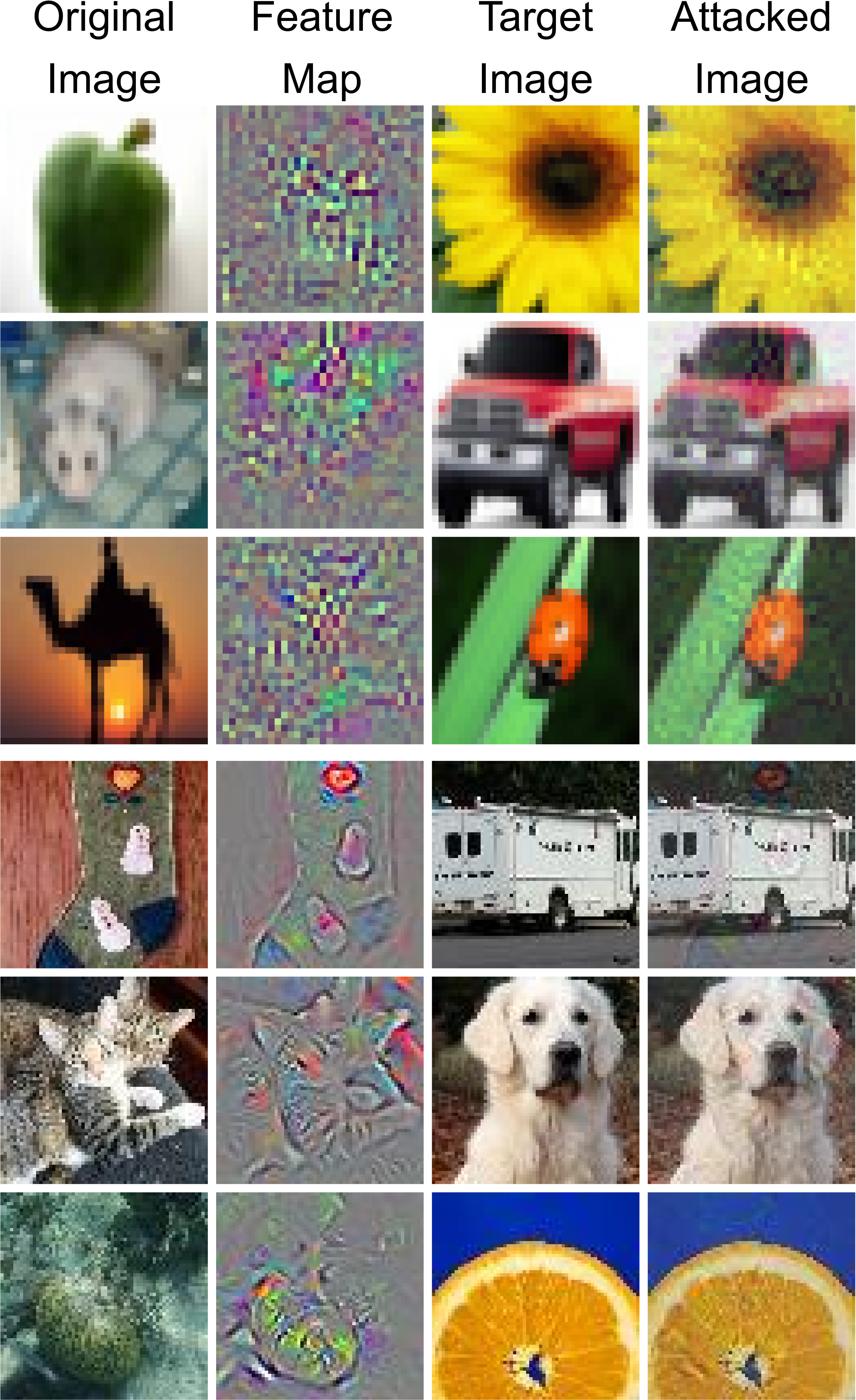}
    \caption{Original Image, Feature Map, Target Image, and Attacked Image for CIFAR, in the first three rows, and Tiny ImageNet, in the last three. The Original Image refers to the image used to obtain the Feature Map, the Target Image is the one whose classification the attacker wants to compromise, and the Attacked Image is the combination of the Feature Map and the Target Image.}
    \label{fig:image_examples}
\end{figure}

\textbf{Qualitative Analysis of ZQBA.}
We provide representative samples for the considered datasets in Figure~\ref{fig:image_examples} to ensure that the proposed attack does not result in significant visual disruption. The examples show that the attacked image does not undergo significant changes relative to its original state, and it is still possible for humans to correctly classify the objects existing in these images. Furthermore, the feature maps obtained for Tiny ImageNet are more concise and clearly represent an object, while CIFAR tends to scatter more around the region of the object, mainly due to the size of the original images. This aspect is also consistent with the results for dataset transferability, in Table~\ref{tab:cross_domain}. These images demonstrate the stealthiness and the applicability of using feature maps to produce adversarial attacks, without needing to query the target model, highlighting the relevance of ZQBA for imperceptible adversarial attacks.

\section{\uppercase{Conclusions}}
\label{sec:conclusions}

This paper proposes ZQBA, a black-box adversarial attack, that can create subtle adversarial perturbations without requiring queries to the target model, by leveraging the representation generated by DNNs to create perturbations that can fool other models (target models). The results show that ZQBA can transfer its adversarial samples across multiple networks and sizes, across various datasets, and achieves better performance than state-of-the-art black-box attacks using a single query, for CIFAR and Tiny ImageNet. We also evaluate the imperceptibility of ZQBA adversarial samples, both quantitatively (SSIM) and qualitatively, showing that these changes do not impair human classification. Finally, ZQBA enables an attacker to generate subtle adversarial samples without querying the target model, highlighting the vulnerabilities of utilizing DNNs in real-world contexts.

\section*{\uppercase{Acknowledgements}}

This work was supported in part by the Portuguese Fundação para a Ciência e Tecnologia (FCT)/Ministério da Ciência, Tecnologia e Ensino Superior (MCTES) through National Funds and co-funded by EU funds under Project UIDB/50008/2020; in part by the FCT Doctoral Grant 2020.09847.BD and Grant 2021.04905.BD.

\bibliographystyle{apalike}

{\small
\begin{thebibliography}{}
\bibitem[Andriushchenko et~al., 2020]{andriushchenko2020square}
Andriushchenko, M., Croce, F., Flammarion, N., and Hein, M. (2020).
\newblock Square attack: a query-efficient black-box adversarial attack via random search.
\newblock In {\em ECCV}, pages 484--501. Springer.

\bibitem[Chen et~al., 2024]{chen2024diffusion}
Chen, J., Chen, H., Chen, K., Zhang, Y., Zou, Z., and Shi, Z. (2024).
\newblock Diffusion models for imperceptible and transferable adversarial attack.
\newblock {\em IEEE Transactions on Pattern Analysis and Machine Intelligence}.

\bibitem[Chen et~al., 2017]{chen2017zoo}
Chen, P.-Y., Zhang, H., Sharma, Y., Yi, J., and Hsieh, C.-J. (2017).
\newblock Zoo: Zeroth order optimization based black-box attacks to deep neural networks without training substitute models.
\newblock In {\em Proceedings of the 10th ACM workshop on artificial intelligence and security}, pages 15--26.

\bibitem[Chen et~al., 2023a]{chen2023advdiffuser}
Chen, X., Gao, X., Zhao, J., Ye, K., and Xu, C.-Z. (2023a).
\newblock Advdiffuser: Natural adversarial example synthesis with diffusion models.
\newblock In {\em Proceedings of the IEEE/CVF International Conference on Computer Vision}, pages 4562--4572.

\bibitem[Chen et~al., 2023b]{chen2023content}
Chen, Z., Li, B., Wu, S., Jiang, K., Ding, S., and Zhang, W. (2023b).
\newblock Content-based unrestricted adversarial attack.
\newblock {\em Advances in Neural Information Processing Systems}, 36:51719--51733.

\bibitem[Costa et~al., 2025]{costa2025lisard}
Costa, J.~C., Roxo, T., Proen{\c{c}}a, H., and In{\'a}cio, P.~R. (2025).
\newblock Lisard: Learning image similarity to defend against gray-box adversarial attacks.
\newblock {\em arXiv preprint arXiv:2502.20562}.

\bibitem[Costa et~al., 2022]{costa2022predicting}
Costa, J.~C., Roxo, T., Sequeiros, J.~B., Proenca, H., and Inacio, P.~R. (2022).
\newblock Predicting cvss metric via description interpretation.
\newblock {\em IEEE Access}, 10:59125--59134.

\bibitem[Croce and Hein, 2020]{croce2020reliable}
Croce, F. and Hein, M. (2020).
\newblock Reliable evaluation of adversarial robustness with an ensemble of diverse parameter-free attacks.
\newblock In {\em ICML}, pages 2206--2216. PMLR.

\bibitem[Dabouei et~al., 2020]{dabouei2020smoothfool}
Dabouei, A., Soleymani, S., Taherkhani, F., Dawson, J., and Nasrabadi, N. (2020).
\newblock Smoothfool: An efficient framework for computing smooth adversarial perturbations.
\newblock In {\em Proceedings of the IEEE/CVF Winter Conference on Applications of Computer Vision}, pages 2665--2674.

\bibitem[Flynn et~al., 2013]{flynn2013image}
Flynn, J.~R., Ward, S., Abich~IV, J., and Poole, D. (2013).
\newblock Image quality assessment using the ssim and the just noticeable difference paradigm.
\newblock In {\em International Conference on Engineering Psychology and Cognitive Ergonomics}, pages 23--30. Springer.

\bibitem[Goodfellow et~al., 2014]{goodfellow2014explaining}
Goodfellow, I.~J., Shlens, J., and Szegedy, C. (2014).
\newblock Explaining and harnessing adversarial examples.
\newblock {\em arXiv preprint arXiv:1412.6572}.

\bibitem[Huang and Tang, 2025]{huang2025scoreadv}
Huang, C. and Tang, H. (2025).
\newblock Scoreadv: Score-based targeted generation of natural adversarial examples via diffusion models.
\newblock {\em arXiv preprint arXiv:2507.06078}.

\bibitem[Imtiaz et~al., 2022]{imtiaz2022saif}
Imtiaz, T., Kohler, M., Miller, J., Wang, Z., Sznaier, M., Camps, O., and Dy, J. (2022).
\newblock Saif: Sparse adversarial and imperceptible attack framework.
\newblock {\em arXiv preprint arXiv:2212.07495}.

\bibitem[Krizhevsky et~al., 2009]{krizhevsky2009learning}
Krizhevsky, A., Hinton, G., et~al. (2009).
\newblock Learning multiple layers of features from tiny images.
\newblock {\em Master's thesis, University of Tront}.

\bibitem[Le and Yang, 2015]{le2015tiny}
Le, Y. and Yang, X. (2015).
\newblock Tiny imagenet visual recognition challenge.
\newblock {\em CS 231N}, 7(7):3.

\bibitem[Liu et~al., 2024]{liu2024difattack}
Liu, J., Zhou, J., Zeng, J., and Tian, J. (2024).
\newblock Difattack: Query-efficient black-box adversarial attack via disentangled feature space.
\newblock In {\em Proceedings of the AAAI Conference on Artificial Intelligence}, volume~38, pages 3666--3674.

\bibitem[Madry et~al., 2018]{madry2018towards}
Madry, A., Makelov, A., Schmidt, L., Tsipras, D., and Vladu, A. (2018).
\newblock Towards deep learning models resistant to adversarial attacks.
\newblock In {\em ICLR}.

\bibitem[Moosavi-Dezfooli et~al., 2016]{moosavi2016deepfool}
Moosavi-Dezfooli, S.-M., Fawzi, A., and Frossard, P. (2016).
\newblock Deepfool: a simple and accurate method to fool deep neural networks.
\newblock In {\em Proceedings of the IEEE conference on computer vision and pattern recognition}, pages 2574--2582.

\bibitem[Mostafa et~al., 2022]{mostafa2022leveraging}
Mostafa, S., Mondal, D., Beck, M.~A., Bidinosti, C.~P., Henry, C.~J., and Stavness, I. (2022).
\newblock Leveraging guided backpropagation to select convolutional neural networks for plant classification.
\newblock {\em Frontiers in Artificial Intelligence}, 5:871162.

\bibitem[Muhammad and Yeasin, 2020]{muhammad2020eigen}
Muhammad, M.~B. and Yeasin, M. (2020).
\newblock Eigen-cam: Class activation map using principal components.
\newblock In {\em 2020 international joint conference on neural networks (IJCNN)}, pages 1--7. IEEE.

\bibitem[Papernot et~al., 2016]{papernot2016limitations}
Papernot, N., McDaniel, P., Jha, S., Fredrikson, M., Celik, Z.~B., and Swami, A. (2016).
\newblock The limitations of deep learning in adversarial settings.
\newblock In {\em 2016 IEEE European symposium on security and privacy (EuroS\&P)}, pages 372--387. IEEE.

\bibitem[Park et~al., 2024]{park2024hard}
Park, J., Miller, P., and McLaughlin, N. (2024).
\newblock Hard-label based small query black-box adversarial attack.
\newblock In {\em Proceedings of the IEEE/CVF WACV}, pages 3986--3995.

\bibitem[Paszke et~al., 2019]{paszke2019pytorch}
Paszke, A., Gross, S., Massa, F., Lerer, A., Bradbury, J., Chanan, G., Killeen, T., Lin, Z., Gimelshein, N., Antiga, L., et~al. (2019).
\newblock Pytorch: An imperative style, high-performance deep learning library.
\newblock {\em Advances in neural information processing systems}, 32.

\bibitem[Patr{\'\i}cio et~al., 2023]{patricio2023coherent}
Patr{\'\i}cio, C., Neves, J.~C., and Teixeira, L.~F. (2023).
\newblock Coherent concept-based explanations in medical image and its application to skin lesion diagnosis.
\newblock In {\em Proceedings of the IEEE/CVF Conference on Computer Vision and Pattern Recognition}, pages 3799--3808.

\bibitem[Roxo et~al., 2024a]{roxo2024asdnb}
Roxo, T., Costa, J.~C., In{\'a}cio, P., and Proen{\c{c}}a, H. (2024a).
\newblock Asdnb: Merging face with body cues for robust active speaker detection.
\newblock {\em arXiv preprint arXiv:2412.08594}.

\bibitem[Roxo et~al., 2024b]{roxo2024bias}
Roxo, T., Costa, J.~C., In{\'a}cio, P.~R., and Proen{\c{c}}a, H. (2024b).
\newblock Bias: A body-based interpretable active speaker approach.
\newblock {\em IEEE Transactions on Biometrics, Behavior, and Identity Science}.

\bibitem[Roy et~al., 2025]{roy2025taigen}
Roy, S., Jain, A., Vatsa, M., and Singh, R. (2025).
\newblock Taigen: Training-free adversarial image generation via diffusion models.
\newblock {\em arXiv preprint arXiv:2508.15020}.

\bibitem[Selvaraju et~al., 2017]{selvaraju2017grad}
Selvaraju, R.~R., Cogswell, M., Das, A., Vedantam, R., Parikh, D., and Batra, D. (2017).
\newblock Grad-cam: Visual explanations from deep networks via gradient-based localization.
\newblock In {\em Proceedings of the IEEE international conference on computer vision}, pages 618--626.

\bibitem[Srinivas and Fleuret, 2019]{srinivas2019full}
Srinivas, S. and Fleuret, F. (2019).
\newblock Full-gradient representation for neural network visualization.
\newblock {\em Advances in neural information processing systems}, 32.

\bibitem[Szegedy et~al., 2014]{szegedy2014intriguing}
Szegedy, C., Zaremba, W., Sutskever, I., Bruna, J., Erhan, D., Goodfellow, I., and Fergus, R. (2014).
\newblock Intriguing properties of neural networks.
\newblock In {\em 2nd International Conference on Learning Representations, ICLR 2014}.

\bibitem[Thirunavukarasu et~al., 2023]{thirunavukarasu2023large}
Thirunavukarasu, A.~J., Ting, D. S.~J., Elangovan, K., Gutierrez, L., Tan, T.~F., and Ting, D. S.~W. (2023).
\newblock Large language models in medicine.
\newblock {\em Nature medicine}, 29(8):1930--1940.

\bibitem[Touvron and et~al., 2023]{touvron2023llama2openfoundation}
Touvron, H. and et~al. (2023).
\newblock Llama 2: Open foundation and fine-tuned chat models.

\bibitem[Wang et~al., 2004]{wang2004image}
Wang, Z., Bovik, A.~C., Sheikh, H.~R., and Simoncelli, E.~P. (2004).
\newblock Image quality assessment: from error visibility to structural similarity.
\newblock {\em IEEE transactions on image processing}, 13(4):600--612.

\end{thebibliography}}

\end{document}